\documentclass[lettersize,journal]{IEEEtran}
\usepackage{amsmath,amsfonts}
\usepackage{algorithmic}
\usepackage{algorithm}
\usepackage{array}
\usepackage{multicol}
\usepackage{textcomp}
\usepackage{stfloats}
\usepackage{url}
\usepackage{verbatim}
\usepackage{graphicx}
\usepackage{cite}
\usepackage{subfigure}
\usepackage{booktabs}
\title{Deep Oscillatory Neural Network}
\author{Nurani Rajagopal Rohan$^{1,2,\dag}$, C. Vigneswaran$^{1,\dag}$, Sayan Ghosh$^{1}$, Kishore Rajendran$^{3}$,  A. Gaurav$^{3}$, V. Srinivasa Chakravarthy$^{1,4,5,*}$

\thanks{$^1$ Computational Neuroscience Lab, Department of Biotechnology, Bhupat and Jyoti Mehta School of Biosciences, Indian Institute of Technology Madras, Chennai, Tamil Nadu, 600036, India

$^2$ The Uncertainty Lab, Department of Applied Mechanics, Indian Institute of Technology Madras, Chennai, Tamil Nadu, 600036, India

$^3$ Department of Electrical Engineering, Indian Institute of Technology Madras, Chennai, Tamil Nadu, 600036, India

$^4$ Department of Medical Science and Technology, Indian Institute of Technology
Madras, Chennai, Tamil Nadu, 600036, India
$^5$ Complex Systems and Dynamics Group, Indian Institute of Technology Madras, Chennai, Tamil Nadu, 600036, India

$^{\dag}$ The first two authors contributed equally to this work

$^*$ Corresponding Author}}

\begin{document}

\maketitle

\begin{abstract}
    We propose a novel, brain-inspired deep neural network model known as the Deep Oscillatory Neural Network (DONN). Deep neural networks like the Recurrent Neural Networks indeed possess sequence processing capabilities but the internal states of the network are not designed to exhibit brain-like oscillatory  activity. With this motivation, the DONN is designed to have oscillatory internal dynamics. Neurons of the DONN are either nonlinear neural oscillators or traditional neurons with sigmoidal or ReLU activation. The neural oscillator used in the model is the Hopf oscillator, with the dynamics described in the complex domain. Input can be presented to the neural oscillator in three possible modes. The sigmoid and ReLU neurons also use complex-valued extensions. All the weight stages are also complex-valued. Training follows the general principle of weight change by minimizing the output error and therefore has an overall resemblance to complex backpropagation. A generalization of DONN to convolutional networks known as the Oscillatory Convolutional Neural Network is also proposed. The two proposed oscillatory networks are applied to a variety of benchmark problems in signal and image/video processing. The performance of the proposed models is either comparable or superior to published results on the same data sets.\\

\textit{Keywords-}complex-valued oscillators, complex-valued weights, brain-inspired networks, sequential problems

\end{abstract}

\begin{section}{Introduction}

There seems to be a deep dichotomy at the heart of the existing approaches to large scale models of brain function. In the recent years, deep learning models, which have not been hitherto taken seriously in terms of biological plausibility, have been shown to accurately capture sensory function in visual and auditory domains \cite{KELL2018630, huang2019braininspired, wardle2020rapid, wardle2020recent, geirhos2022imagenettrained, cichy2016comparison}. However, these models have not been able to describe the internal dynamics of the brain involved in sensory processing. On the other hand, networks of nonlinear oscillators have been shown to accurately model brain dynamics as described by functional neuroimaging methods \cite{bick2020understanding, chinichian2024modeling, laing2017phase}. However, these oscillator networks, in their current form, are unable to learn input/output behaviour. Thus, there is need to develop a class of models that resolve this dichotomy and serve to capture both input/output behaviour and brain dynamics. Let us now take a closer look at this dichotomy and explore a path to its resolution.

Recurrent Neural Networks (RNNs) are traditionally used to solve sequential tasks in a variety of domains ranging from Natural Language Processing (NLP), time-series forecasting, music generation, video analysis, etc. Popular variants of vanilla RNNs like Long Short-Term Memory units (LSTMs), Gated Recurrent Unit (GRUs), flip-flop neural networks are widely used for different kinds of sequential problems and achieve state-of-the-art performances \cite{flipflop}. These sequential networks can operate in different modes namely, many-to-one (eg. sentiment analysis) \cite{gopalakrishnan2020sentiment}, one-to-many (eg. handwriting generation) \cite{graves2014generating} and, many-to-many (eg. machine translation) \cite{TAN20205}. Recurrent units with convolutional operations on gating variables are used to capture spatio-temporal features in video processing tasks \cite{shi2015convolutional}. In the recent years, deep neural networks (DNN) have been shown to match human performance in pattern recognition in visual and auditory domains \cite{KELL2018630, huang2019braininspired}. Studies that demonstrated a similarity in human performance with that of DNNs, have further shown that even human error pattern with respect shape variation \cite{atabay2017deep}, viewpoint variation \cite{wardle2020rapid, wardle2020recent} and object class \cite{cichy2016comparison, geirhos2022imagenettrained} also matched with DNN error patterns. However, this broad match in error patterns did not carry over into errors committed on classification of individual images \cite{rajalingam2018review}. Similarly in the auditory domain, studies based on DNNs have shown a close performance match in music and speech recognition \cite{music_ref, hinton2012deep, chou2019learning}. Despite these remarkable successes in achieving human-like behavioral performance, the aforementioned deep learning models are unable to capture the oscillatory activity of brain dynamics.

There is a long history of studies that link oscillatory activity in the brain with cognitive functions \cite{Kahana1669} and brain states in normal (eg. sleep and waking) \cite{dang2020neuronal} and abnormal conditions (eg. anesthesia \cite{akeju2017neural} and epilepsy \cite{blanco2011data}). Such studies have resulted in an understanding that perception, memory and other higher cognitive functions in the brain are implemented by synchronized oscillatory networks in the brain. Oscillatory neural activity in the brain is thought to span a frequency range of 0.05 to 500 Hz \cite{doi:10.1126/science.1099745}. This broad range of frequencies has been divided into about a dozen frequency bands which show a linear progression in the logarithmic scale. Each of these frequency bands have been associated with specific brain states with neighbouring bands linked to similar states. 

Considering the importance of oscillations in brain function, it is desirable that any large-scale model of brain function must necessarily accommodate this oscillatory brain activity and echo the empirically determined significance of various brain frequencies. Neural mass models \cite{pinotsis2014neural} have been used effectively to describe brain dynamics at mesoscopic and macroscopic levels. In these models the fundamental unit is not a single neuron but a neural ensemble, whose collective activity is described by nonlinear “neural” oscillators. Popular low- dimensional oscillator models like the Hopf oscillator, Wilson-Cowan oscillator \cite{cowan1972autoradiographic}, FitzHugh-Nagumo \cite{nagumo1962active} oscillator and Kuramoto oscillator have been used in these networks. Such oscillatory networks have been used to describe brain dynamics as expressed through functional imaging techniques like fMRI, MEG and high-density EEG \cite{jirsa2022virtual, bandyopadhyay2023phenomenological, cabral2011role, bey2023lesion, sanz2013virtual, deco2009key}. However, though these oscillatory networks are able to capture brain dynamics, they are unable to learn input-output behaviour. Furthermore, in those instances where oscillatory networks have been used describe behaviour, they are typically restricted to intrinsically oscillatory or rhythmic behaviours like locomotive movements \cite{dutra2003modeling}, rhythmic hand movements \cite{lakatos2014nonlinear}, swimming movements \cite{cao2013applying} etc.

From the above considerations, there is a clear need to develop deep neural networks in which the hidden layers are constituted of oscillatory neuron models. The networks must be capable of learning input/output behaviour like any deep neural network. In this paper we address this challenge and present a general class of trainable deep oscillatory neural networks. 

The outline of the paper is as follows. Section 2 describes the model development and equations that govern dynamics and learning. In Section 3, we apply this network to a class of benchmark problems that involve sequential processing. We describe the results obtained and compare them with results from other non-oscillatory deep neural networks. Section 4 discusses the results and outlines how the present class of networks can be developed further to improve their biological plausibility.
    
\end{section}
\begin{section}{Methods}

    The oscillatory neuron model we use in our networks is the Hopf oscillator. It is a harmonic oscillator with a stable limit cycle. The canonical Hopf oscillator is described by the following complex valued differential equation, 
    \begin{equation}
        \dot{z} = z\left(\mu + \iota\omega + \beta_1|z|^2 + \frac{\epsilon\beta_2|z|^4}{1 - \epsilon|z|^2} \right) + I(t).
    \end{equation}
    where $\omega$ is the natural angular frequency of the Hopf oscillator, $I(t)$ is the external input to the oscillator. The polar coordinate representation is more explicate for our analysis is given as follows,
    \begin{align}
            \dot{r} &= \mu r + \beta r^3 + \frac{\epsilon \beta_2 r^5}{1 - \epsilon r^2} + A(t)~ \cos ~\psi \nonumber \\
            \dot{\psi} &= \Omega - \frac{A(t)}{r}~ \sin~ \psi
            \label{eq:canonical_hopf}
        \end{align}
    where $I(t) = A(t)e^{i\phi(t)}$, $\psi = \theta - \phi(t)$ and $\Omega = \omega - \frac{\mathrm{d}\phi}{dt}$ is the difference between the angular frequencies of the oscillator and the external input. The oscillator exhibits four distinct dynamical behaviours characterised by the parameters $(\mu, \beta_1, \beta_2)$ $-$ critical Hopf regime $(\mu = 0, \beta_1 < 0, \beta_2 = 0)$, supercritical Hopf regime $(\mu > 0, \beta_1 < 0, \beta_2 = 0)$, double limit cycle regime $(\mu < 0, \beta_1 > 0, \beta_2 < 0)$. We restrict our attention to the critical or supercritical regime; for a more detailed analysis on the others see \cite{kimandlarge}. The equations then transform to, 
    \begin{equation}
        \dot{z} = z\left(\mu + \iota\omega + \beta|z|^2 \right) + I(t).
    \end{equation}
    \begin{align}
        \dot{r} &= \mu r + \beta r^3 + A(t)~ \cos ~\psi \nonumber \\
            \dot{\theta} &= \Omega - \frac{A(t)}{r}~ \sin~ \psi
        \label{eq:canonical_hopf_without_fifth_order_term}
    \end{align}
    
    The proposed oscillator neural network model consists of successive layers of static dense or convolutional layers with nonlinear activation and dynamic Hopf oscillator layers. A general oscillator neural network is depicted in Fig. \ref{fig:general_ONN}. There are three modes in which an input, $z_{in}(t)$ can be presented to the oscillator layer for forward propagation; either as external input $I(t)$, referred to as resonator mode, or as input to the amplitude $\mu(t)$, commonly referred as amplitude modulation, or as input to the frequency $\omega(t)$, referred to as frequency modulation (fig. 2). The method of forward pass for the three modes of input to a single oscillator unit is described below, 
    
    \begin{figure}[ht]
        \centering
        \includegraphics[height = 5cm, width = 8cm]{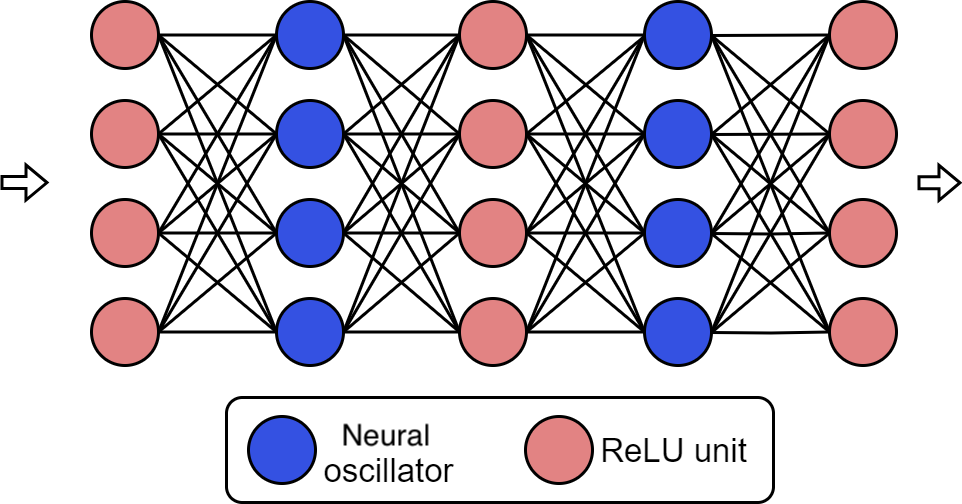}
        \caption{The Oscillatory Neural Network is generally composed of alternating static and oscillatory layers.}
        \label{fig:general_ONN}
    \end{figure}

        \begin{figure}
        \centering
        \includegraphics[height = 4cm, width = 7cm]{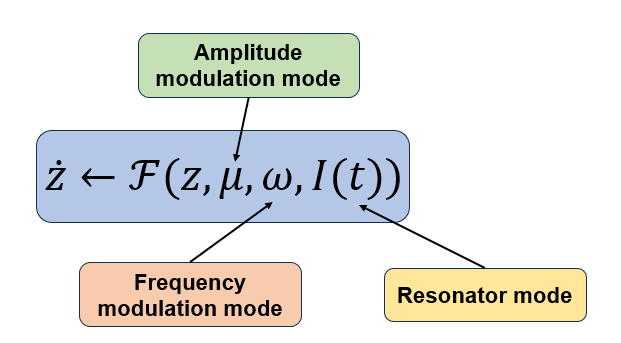}
        \caption{Mechanisms through which the input, $z_{in}(t)$ can be presented to the Oscillator unit}
        \label{fig:various_input_modes}
    \end{figure}

    \begin{figure}[h]
        \centering
        \includegraphics[height = 8cm, width = 8cm]{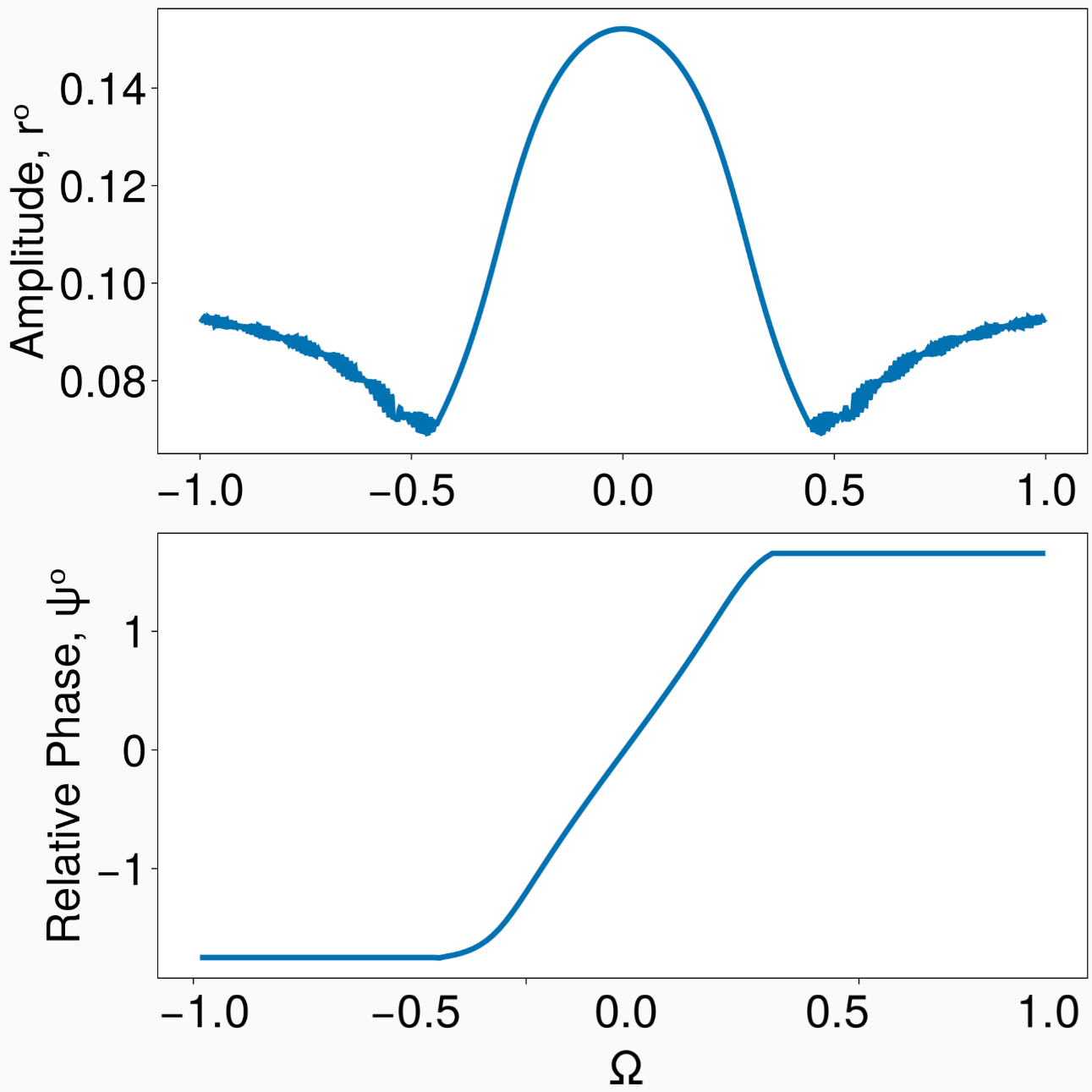}
        \caption{Driven Behaviour of a Supercritical Hopf Oscillator Under Strong Forcing. Steady-state amplitude and relative phase as a function of frequency difference $(\mu = 1, \beta_1 = -100, \beta_2 = 0, F = 0.2).$}
        \label{fig:Resonant Tuning Curve}
    \end{figure}

    \begin{enumerate}
        \item \textbf{Resonator mode (Input presented as $I(t)$):}
        In the presence of an external complex periodic driving of angular frequency $\omega_0$, $z_{in}(t) = I(t) = I_0 e^{\iota\omega_0t}$, Eq. \ref{eq:canonical_hopf_without_fifth_order_term} for the case of supercritical Hopf bifurcation transforms to,
        \begin{align}\label{eq:canonical_hopf_with_forcing}
            \dot{r} &= \mu r + \beta r^3  + \kappa_I I_0~ \cos ~\psi \nonumber \\
            \dot{\theta} &= \Omega - \kappa_I\frac{I_0}{r}~ \sin~ \psi
        \end{align}
        where $\psi = \theta - \omega_0t$ and $\Omega = \omega - \omega_0$ is the difference between the angular frequencies of the oscillator and the external input. In the presence of strong forcing $(I_0 \approx 10^{-1})$, the behaviour of a supercritical Hopf oscillator is given in Fig. \ref{fig:Resonant Tuning Curve}. The oscillator shows resonance in a small range of frequencies around its natural frequency. The oscillator activation is then in a frequency locked state with the input signal i.e., the frequency of the oscillator activation is equal to the frequency of the input signal. Furthermore, within the smaller range of frequency, the oscillator activation is in a phase locked state to the input signal (Fig \ref{fig:Resonant Tuning Curve}). Beyond this regime, the system undergoes phase slip. We note that the resonance described in Fig. \ref{fig:Resonant Tuning Curve} is only valid for when the input is of the particular form, $I(t) = I_0 e^{\iota\omega_0t}$.

        \item \textbf{Amplitude Modulation (Input presented as $\mu(t)$):}
        In this case, with $z_{in}(t)$ as input to the $r$ equation of the oscillator, the dynamical equations are given by, 
        \begin{align}
            \mu(t) &= (\mu_0 + \kappa_{\mu} \Re (z_{in}(t)))\\
            \dot{r_i} &= \mu(t)r_i + \beta r_i^3  \nonumber \\
            \dot{\theta_i} &= \omega_i
            \label{eq:canonical_hopf_with_mu_feedforward}
        \end{align}
        where, $\Re$ denotes the real part of the complex number. Since the amplitude of the oscillator is a positive quantity, one must ensure that, $\kappa_{\mu} \mathrm{min}(\Re (z_{in})) + \mu_0 > 0$ where $\mathrm{min}(\Re (z_{in}))$ is the minimum value of $\Re(z_{in}(t))$.

        \item \textbf{Frequency Modulation (Input presented as $\omega(t)$):}
        In this case, $z_{in}(t)$ is given as input to the $\theta$ equation of the oscillator. In this case, the input modulates the effective oscillator frequency as follows: 
        \begin{align}
            \dot{r_i} &= \mu r_i + \beta r_i^3  \nonumber \\
            \dot{\theta_i} &= \omega_i + \kappa_{\omega}\Re(z_i(t))
            \label{eq:canonical_hopf_with_omega_feedforward}
        \end{align}
    \end{enumerate}
    In the latter two modes of the input, the Hopf oscillation activation gets modulated by the input oscillations, hence the name. In this paper, we only use the first two modes of input presentation to the neural oscillator. The above differential equations are solved from $t = 0 \ldots T$ using Forward Euler method with fixed time step relevant to the dataset. Suppose the solution to either of the above system is given $\{r, \theta\}$, then the output is given by,
    \begin{align}
        \mathfrak{Z}_{d}(t) &= (r \cos \theta + \iota ~ r \sin \theta)          \label{eq:canonical_hopf_with_forcing_feedforward_output}
    \end{align}

    As depicted in Fig. \ref{fig:general_ONN}, the complex activation (Eq. \ref{eq:canonical_hopf_with_forcing_feedforward_output}) of the oscillator layer feeds into a static layer. The complex nonlinear activation, $\mathfrak{f}$ is then applied as follows, 
    \begin{align}
        \mathfrak{Z}_s(t) &= \mathfrak{f}(\Re(\mathfrak{Z}_d(t))) + \iota  \mathfrak{f}(\Im(\mathfrak{Z}_d(t)))
    \end{align}
    point wise in time. Where $\Re,  \Im$ denote the real part and imaginary part of the complex numbers. The complex nonlinear activation $\mathfrak{f}$ can be ReLU, $\tanh$, or sigmoid function.\\

    All the functions mentioned above are continuously differentiable. We use automatic differentiating feature (autograd) of TensorFlow library to calculate the gradients for backpropogation to update network weights. 

    \subsection{Oscillatory Convolutional Neural Network (OCNN)}
    In our proposed OCNN model, due to the presence of oscillatory elements, neurons of higher layers possess spatio-temporal responses. Arrangement of oscillators are identical to the input feature maps to maintain the spatial information. The intrinsic frequencies of the oscillators are sampled from a uniform random distribution of appropriate range. The frequency distribution is passed through a Gaussian blur of kernel size 3x3 across the 2D arrangements of oscillators in each channel. Hierarchical arrangements of such pairs of convolution and oscillators enable learning and detection of features at different levels of complexities in spatiotemporal scale, see Fig. \ref{fig:OCNN_block}.

        \begin{figure}[h]
        \centering
        \includegraphics[height = 5cm, width = 8cm]{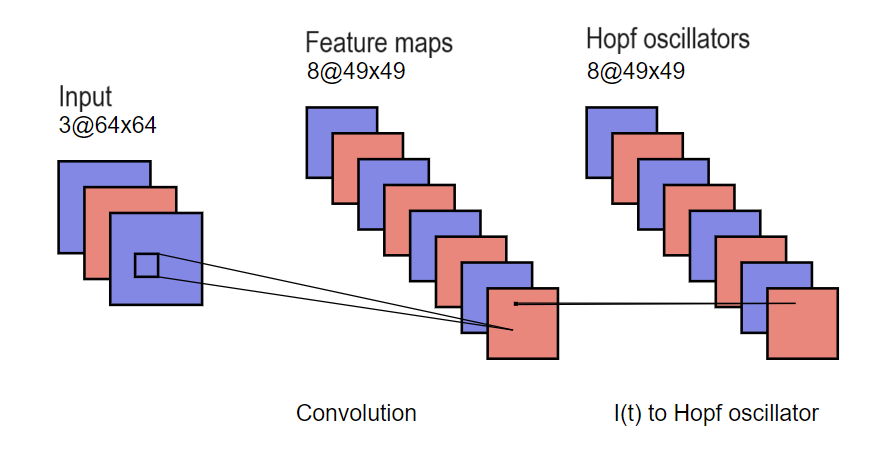}
        \caption{Convolutional Oscillatory block: feature maps obtained after convolutional operation on input image at time $t$ is given as input $I(t)$ in one-to-one fashion to oscillatory block of dimension equal to that of feature maps to obtain spatio-temporal feature maps}
        \label{fig:OCNN_block}
        \end{figure}

    \subsection{Classification using Oscillatory Neural Network}
    Due to the dynamic nature of the DONN model, the output of the network is always a time series. If the network is tasked with a classification problem, the labels have to be defined differently.  In order to have an equivalent classification score for oscillator activation, we use the ramp approach to decide the predicted class. Ramp-like classifications are often used in biological models as they closely replicate the race-based threshold criterion to select the winner neurons \cite{saccade}. For each time step, all classes’ (corresponding neurons in the last layer) desired predictions are zero, except for the desired class where the desired value rises linearly. The predicted class is decided by selecting the neuron which has the maximum average across time steps.
\end{section}

\begin{section}{Experiments and Results}
    The DONN was implemented in Python 3.9.11 and TensorFlow 2.10.1, and on an x86-CPU and NVIDIA GPU running windows 11. We demonstrate the capability of DONN in a variety of basic sequential processing tasks. 
    \subsection{Signal Generation}
        We begin with a simple demonstration of signal generation with DONN. The network is trained to produce a sinusoidal signal of a specific frequency when prompted by a specific label presented as input in 1-hot form. The labels are [1,2,3,4] and and the corresponding signal has frequency [1, 5, 7, 9] in Hertz. The size details of the network architecture are described in Table-\ref{table:signal_generation}.
        \begin{table}[h]
            \centering
            \caption{Network Architecture for Signal Generation task}
            \begin{tabular}{@{}cc@{}}
\toprule
Input frequency range                  & {[}1-9 Hz{]}                                                                                                                          \\ \midrule
Initial frequency range of oscillators & {[}1-10 Hz{]}                                                                                                                         \\ \midrule
Architecture                           & \begin{tabular}[c]{@{}c@{}}  ReLU (10),  Hopf (10),\\  $\tanh$ (10)\\  output (1)\end{tabular} \\ \midrule
Input type to oscillators              & $\mu(t)$                                                                                                                                  \\ \midrule
Frequency of oscillators               & Not trained                                                                                                                           \\ \bottomrule
\end{tabular}
            \label{table:signal_generation}
        \end{table}

        \begin{figure}[h]
            \centering
          \includegraphics[height = 6cm, width = 7cm]{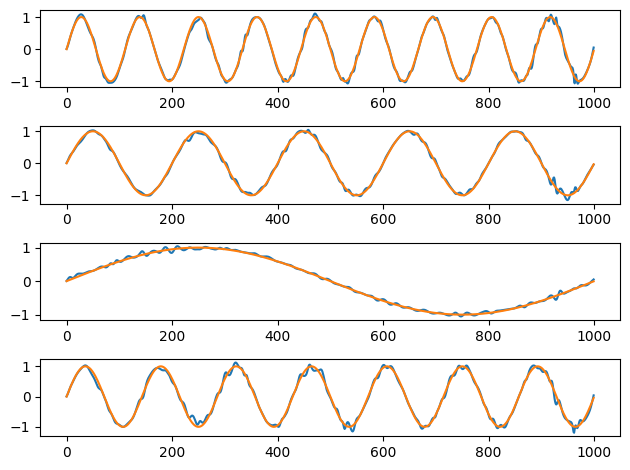}
          \caption{The performance of the model for signal generation task to generate the four classes reliably.}
          \label{fig:signal_generation}
        \end{figure}

        For this problem, the input is presented as $\mu(t)$. The network is able to generate the signals accurately, see Fig. \ref{fig:signal_generation}. The validation Mean Squared Error (MSE) loss is $0.01$ (p$<$0.05, n=10).
        
    \subsection{Amplitude Demodulation}
        In order to transmit a signal over a distance without being affected by external interference or noise and without experiencing degradation, it must undergo a modulation process. The inverse process of retrieving the message is demodulation. We show that our network is able to perform this operation. The mathematical expression for amplitude modulation is given as follows, 
        \begin{align}
            \mathcal{M}(t) = (1 + m(t)) \sin(\omega_c t)
        \end{align}
        where $m(t)$ is the message signal, $\omega_c$ is the carrier wave frequency and M(t) is the modulated signal. For this task the message signal, $m(t) = \sum_{i = 1}^5 \sin(2\pi f_i t)$, $f_i \in \mathcal{U}(1, 5)$ Hz. Carrier frequency, $\omega_c = 8$ Hz is kept fixed. Some sample message signals and the corresponding modulated signals are given in Fig. \ref{fig:sample_modulated}. The size details of the network architecture are described in Table-\ref{table:amplitude_demodulation}. The oscillators are initialised with frequency in the range 0.1 Hz to 12 Hz. The input to the oscillator is given as $I(t)$. The network is able to demodulate the message signal accurately (fig. \ref{fig:sample_demodulated}). The validation MSE loss is $0.02$ (p$<$0.05, n=10).

        \begin{table}[h]
            \caption{Network architecture for Amplitude Demodulation}
            \centering
            \begin{tabular}{@{}ll@{}}
\toprule
Input frequency range                  & Carrier: 8 Hz, message signal: {[}1-5 Hz{]}                                                                                           \\ \midrule
Initial frequency range of oscillators & {[}0.1-12 Hz{]}                                                                                                                       \\ \midrule
Architecture                           & \begin{tabular}[c]{@{}l@{}} ReLU (40),  Hopf (40),\\  ReLU (40),  Hopf (40),\\  $\tanh$ (40), \\ output (1)\end{tabular} \\ \midrule
Input type to oscillators              & I(t)                                                                                                                                  \\ \midrule
Frequency of oscillators               & Not trained                                                                                                                           \\ \bottomrule
\end{tabular}
            \label{table:amplitude_demodulation}
        \end{table}
                
        \begin{figure}[h]
            \centering
            \begin{subfigure}{}
                \centering
                \includegraphics[height = 4cm, width=5.5cm]{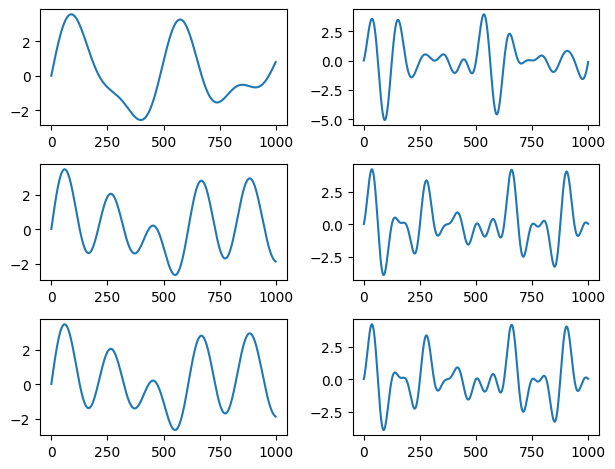}
                \caption{(left) The message signals composed of up to five frequency components, and (right) the corresponding modulated signals.}
                \label{fig:sample_modulated}
            \end{subfigure}%
            \hspace{1em}%
            \begin{subfigure}{}
                \centering
                \includegraphics[height = 4cm, width=6cm]{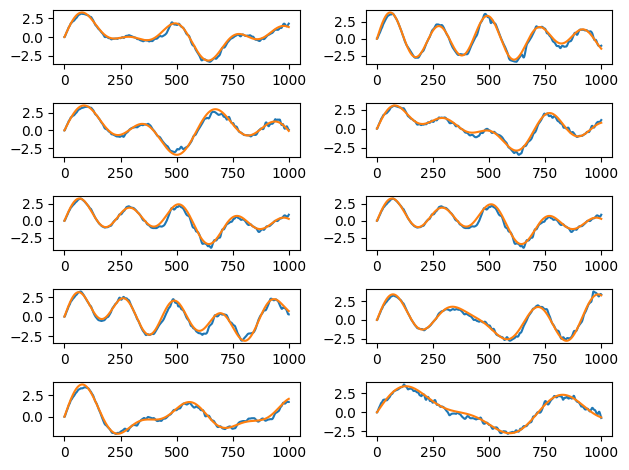}
                \caption{The performance of our model in demodulating the signal. The desired demodulated signal is \textit{blue} and the model predictions are in \textit{orange}.}
                \label{fig:sample_demodulated}
            \end{subfigure}%
        \end{figure}

    \subsection{Signal Filtering}
        In a variety of signal processing problems there are requirements to remove, suppress and/or amplify certain frequency components. Here we apply our model to learn the transfer function of a band-pass filter. 
        
        The input data is composed of synthetically generated sinusoidal signals with frequency in the range 5Hz to 300Hz, and arbitrary phase delays and amplitudes. The signal is generated for 1 second with a sampling rate, $f_s = 1000$. Gaussian noise, $\mathcal{N}(0, 0.1)$, is added to the signal. The signal was filtered using Infinite Impulse Response Butterworth filter of order 4 using SciPy library, allowing frequencies in the range 50Hz to 100 Hz to pass and attenuates others. Some sample input and the corresponding desired filtered output signals are shown in Fig. \ref{fig:sample_io}. 

        The size details of the network architecture are described in Table-\ref{table:frequency_transformation}. The loss function is the Mean Squared Error (MSE), and Adaptive Moment Estimation (ADAM) is used to optimise the weights. The validation MSE loss is $0.004$ (p$<$0.01, n=10). Fig. \ref{fig:tf} shows the learnt transfer function and Fig. \ref{fig:outout} shows the desired and predicted model output.

        \begin{table}[h]
            \centering
            \caption{Network Architecture for Signal Filtering}
            \begin{tabular}{@{}cc@{}}
\toprule
Input frequency range                  & {[}5-300 Hz{]}                                                                                                \\ \midrule
Initial frequency range of oscillators & {[}30-150 Hz{]}                                                                                               \\ \midrule
Architecture                           & \begin{tabular}[c]{@{}c@{}}  ReLU (40),  Hopf (40),\\  $\tanh$ (40),\\  output (1)\end{tabular} \\ \midrule
Frequency of oscillators               & Trained                                                                                                                           \\ \bottomrule
\end{tabular}
            \label{table:frequency_transformation}
        \end{table}

        \begin{figure}[h]
            \centering
            
              \includegraphics[width=6cm]{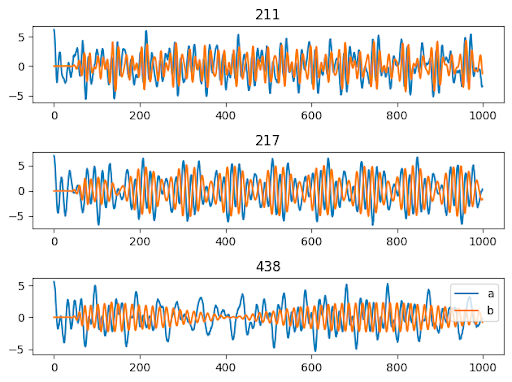}
                  \caption{The figure shows some sample input (Blue) and output signals (Orange)}
                  \label{fig:sample_io}

        \end{figure}       
        \begin{figure}
              \centering
                \begin{subfigure}{}
                    \includegraphics[height=5cm, width=5.5cm]{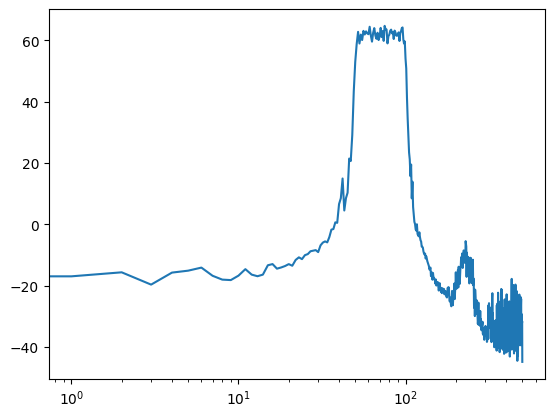}
                    \caption{The figure shows the predicted transfer function of the dataset}
                    \label{fig:tf}
                \end{subfigure}%
                
                \begin{subfigure}{}
                  \includegraphics[width=6cm]{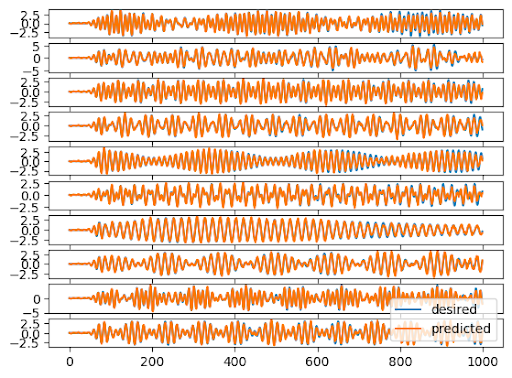}
                  \caption{The figure shows the desired input and output behaviour expected from our network. The predicted signal (Orange) after filtering; upon comparing with the desired signals, we infer that the model is able to capture the Transfer function of the desired Band-Pass filter.}
                  \label{fig:outout}
                \end{subfigure}
            \label{fig:Filter}
        \end{figure}

        \subsection{Learning Mathematical Operators}
            In this task the network is trained to perform indefinite integration and differentiation on synthetically generated sinusoidal signals. Task-1, the data is composed of signals with multiple frequency components and arbitrary initial phase offset and amplitude. If $I(t) = \sum_{i} a_i \sin(\omega_i t + \phi_i)$ is the input signal, the output signal, then for the integration task, $O(t)$ is given by, 
            \begin{equation}
                O(t) = \sum_{i} -\frac{a_i}{\omega_i}\cos(\omega_i t + \phi_i)
            \end{equation}
            and for the differentiation is given by, 
            \begin{equation}
                O(t) = \sum_{i} a_i\omega_i \cos(\omega_i t + \phi_i)
            \end{equation}
            where $a_i, \phi_i, \omega_i$ are the amplitude, phase offsets and the angular frequencies sampled from  $\mathcal{N}(0, 1), ~\mathcal{N}(0, \pi)$ and $\mathcal{U}(1, 5)$ respectively. Task-2, We also apply the model on a data set of composed of trajectories where an agent is allowed to move in a bounded domain following an arbitrary velocity profile as described in \cite{soman}. The network is tasked to perform integration of the velocity profiles, ($\dot{x}, \dot{y}$) to produce the trajectory, $(x, y),$ see Fig. \ref{fig:sample_trajectory_integration}.
            
            \begin{figure}[h]
            \centering
                \begin{subfigure}{}
                    \centering
                    \includegraphics[height = 6cm, width = 5cm]{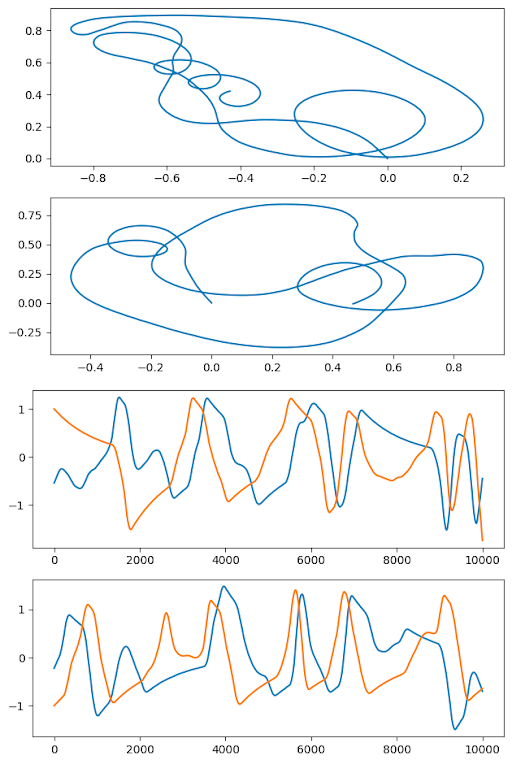}
                    \caption{Sample trajectory (above) and the corresponding velocity profiles (velocities along x and y axis) (below).}
                    \label{fig:sample_trajectory_integration}  
                \end{subfigure}%
                \begin{subfigure}
                    \centering
                    \includegraphics[height = 6cm, width = 7cm]{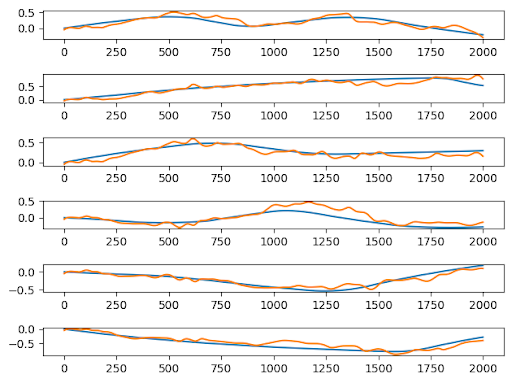}
                    \caption{The model is able to integrate the signals. (Blue) represents the desired and (orange) represents the model predicted output. }
                    \label{fig:trajectory_velocity}
                \end{subfigure}

            \end{figure}
            
            \begin{figure}
                \begin{subfigure}{}
                    \centering
                    \includegraphics[height = 6cm, width = 7cm]{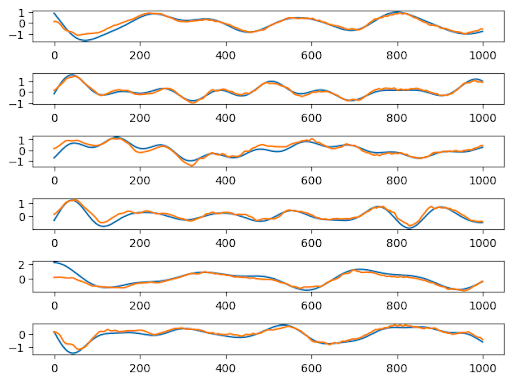}
                    \caption{The network is able to learn the Integration operation, some sample figures are shown. Desired and model predicted output are shown in blue and orange respectively.}
                    \label{fig:integration}
                \end{subfigure}
                \begin{subfigure}{}
                    \centering
                    \includegraphics[height = 6cm, width = 7cm]{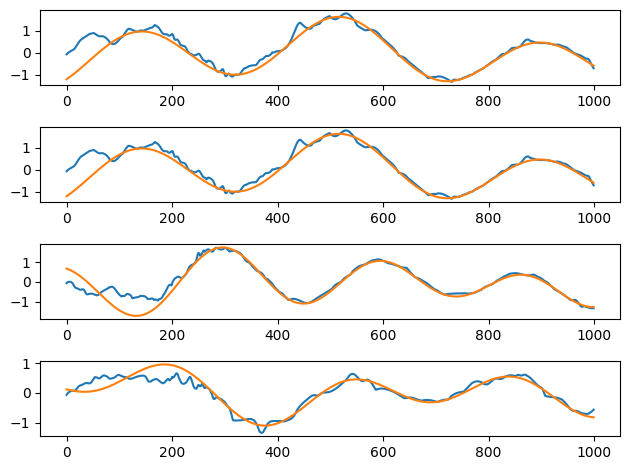}
                    \caption{The network is able to learn the Differentiation operation, some sample figures are shown. Desired output (orange) and model predicted output (blue) are shown.}
                    \label{fig:differentiation}
                \end{subfigure}
            \end{figure}

            The size details of the network architecture for the two tasks are described in Table-\ref{table:integration}.  The network successfully performs the integration operation on both the data sets and the differentiation operation on the former dataset, see Fig. \ref{fig:trajectory_velocity}, \ref{fig:integration}, \ref{fig:differentiation}. The MSE loss for Differentiation is 0.1. For integration, Task-1 the validation MSE loss is 0.08 (p$<$0.05, n=10), Task-2 validation MSE loss is 0.07 (p$<$0.05, n=10).

\begin{table}[h]
\centering
\caption{Network architecture for Learning Mathematical Operations}
\begin{tabular}{@{}ccc@{}}
\toprule
                                       & Task 1                                                                                                                                   & Task 2                                                                                                                                   \\ \midrule
Input frequency range                  & {[}0.1-5 Hz{]}                                                                                                                           & \begin{tabular}[c]{@{}c@{}}Layer-1 : {[}0.1-10 Hz{]}\\ Layer-2 : {[}0.1-2 Hz{]}\end{tabular}                                             \\ \midrule
Initial frequency range \\ of oscillators & {[}1-10 Hz{]}                                                                                                                            & {[}1-10 Hz{]}                                                                                                                            \\ \midrule
Architecture                           & \begin{tabular}[c]{@{}c@{}} ReLU (20),  Hopf (20),\\ ReLU (20), Hopf (20),\\  $\tanh$ (20), \\  output (1)\end{tabular} & \begin{tabular}[c]{@{}c@{}} ReLU (20),  Hopf (20),\\  ReLU (20),  Hopf (20),\\  $\tanh$ (20), \\  output (1)\end{tabular} \\ \midrule
Input type to oscillators              & I(t)                                                                                                                                     & I(t)                                                                                                                                     \\ \midrule
Frequency of oscillators               & Not trained                                                                                                                              & Trained                                                                                                                                  \\ \bottomrule
\end{tabular}
\label{table:integration}
\end{table}

        \subsection{Sentiment Analysis}
        
Movie review sentiment analysis typically involves a two-class classification problem (positive/negative). The IMDB large movie review dataset \cite{maas-EtAl:2011:ACL-HLT2011} serves as a common benchmark for sentiment analysis tasks, with a predefined maximum review length of 500 words. The training data is split into training and validation in a 7 : 3 ratio. The selection of this task aims to demonstrate the proficiency of the Deep Oscillator Neural Network in sequence classification. 

\begin{table}[h]
\centering
\caption{Sentiment Analysis}
\begin{tabular}{@{}ccc@{}}
\toprule
Model                    & Validation accuracy & Architecture                                                                                                                                                                                                                                                                                              \\ \midrule
\begin{tabular}{cc}
Bidirectional \\ LSTM  \cite{flipflop}
\end{tabular}     & 85.19\%             & \begin{tabular}[c]{@{}c@{}} Embedding layer (100), \\ 2 x Bidirectional flip-flops (100),\\  $\tanh$ (20),  output (2)\end{tabular}                                                                                                                                                                 \\ \midrule
\begin{tabular}{cc}
Bidirectional \\ flipflop  \cite{flipflop}
\end{tabular} & 85.07\%             & \begin{tabular}[c]{@{}c@{}} Embedding layer (100), \\ 2 x Bidirectional flip-flops (100),\\  $\tanh$ (20),  output (2)\end{tabular}                                                                                                                                                                 \\ \midrule
\textbf{DONN}                     & 85.2 \%              & \begin{tabular}[c]{@{}c@{}} Embedding layer (100), \\  Hopf (100),  ReLU (100),\\  Hopf (100),  ReLU (100),\\  $\tanh$ (20), \\ output (2)\\ \\ Initial frequency range of \\oscillators: {[}1-15 Hz{]}, \\ Input type to oscillators: I(t), \\ Frequency of \\oscillators: trained\end{tabular} \\ \bottomrule
\end{tabular}
\label{table:SA}
\end{table}

        The words are encoded and passed to the embedding layer followed by the DONN. The embedding dimension is 100 with a vocabulary length of 35,000. The input to the oscillator layer is presented as $I(t)$. The total trainable parameters is 26,798. The optimiser used is the ADAM optimiser with learning rate = 0.001, and the objective function used is mean squared error. The model achieves a performance of 85.2\% (p$<$0.05, n=10) accuracy on testing data. The network architecture and comparison with other models are described in Table-\ref{table:SA}

        \subsection{Action Recognition : UCF 11}
        To showcase the sequence processing capacity of OCNN network in spatiotemporal domain, we validate the network using UCF11 YouTube Action dataset\cite{ucf11_dataset}. UCF11 is a standard 3-channel RGB video classification dataset with different viewpoints, backgrounds and camera motions. The dataset consists of 11 action classes, and number of frames for all the videos is set to 50, with each frame resized to 48$\times$48 from the original 224$\times$224, on account of limited computational capability. Dataset is divided into training and validation with 1290 and 305 samples respectively. 

        The network used for the task consists of two pairs of convolutional-oscillators layers, followed by flattening and dense layers. The architecture details of the network used for the task is mentioned in Table-\ref{table:action_recognition}. Time step of the oscillators is set to 0.02 seconds with the number of time steps equal to the number of frames in a video sample. Adam (learning rate: 0.0001) and MSE are used as optimizer and loss function respectively for training.
        
        \begin{table}[h]
        \centering
            \caption{Action Recognition}
            \begin{tabular}{@{}ccc@{}}
\toprule
Model                                                                          & Validation accuracy & Architecture                                                                                                                                                                                         \\ \midrule
\begin{tabular}[c]{@{}c@{}}Convolutional \\LSTM  \cite{flipflop}\end{tabular}       & 95.42\%             & \begin{tabular}[c]{@{}c@{}}2 x ConvLSTM (3x3,40), \\  Conv3D (3x3x3, 1), \\ flatten, \\  softmax (2)\end{tabular}                                                               \\ \midrule
\begin{tabular}[c]{@{}c@{}}Convolutional \\flip-flops \cite{flipflop} \end{tabular}   & 99.75\%             & \begin{tabular}[c]{@{}c@{}}2 x ConvLSTM (3x3,40), \\  Conv3D (3x3x3, 1), \\ flatten, \\  softmax (2)\end{tabular}                                                               \\ \midrule
\begin{tabular}[c]{@{}c@{}}\textbf{OCNN}\end{tabular} & 98.64\%             & \begin{tabular}[c]{@{}c@{}}2 x OCNN (3x3,40), \\ flatten, \\  output (2)\\ \\ Initial frequency range: {[}1-15 Hz{]},\\Input type to oscillator: I(t), \\ Frequency of oscillators: trained\end{tabular} \\ \bottomrule
\end{tabular}
            \label{table:action_recognition}
        \end{table}

         Fig. \ref{fig:ucf_acc} shows the accuracy and loss curves on validation data, and Fig. \ref{fig:UCF} shows the desired output and output from the model. The validation MSE loss and accuracy of the model is 0.0564 and 99.75\% (p$<$0.05, n=10) respectively. 

       \begin{figure}[h]
           \includegraphics[height = 6cm, width = 8cm]{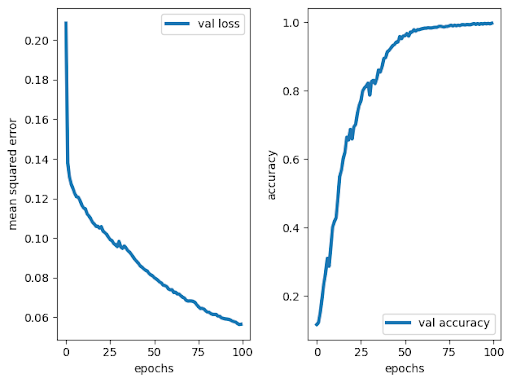}
           \caption{ Accuracy and loss plots on validation data for UCF11 dataset}
            \label{fig:ucf_acc}
        \end{figure}
        
        \begin{figure}
            \centering
            \includegraphics[height = 6cm, width = 8cm]{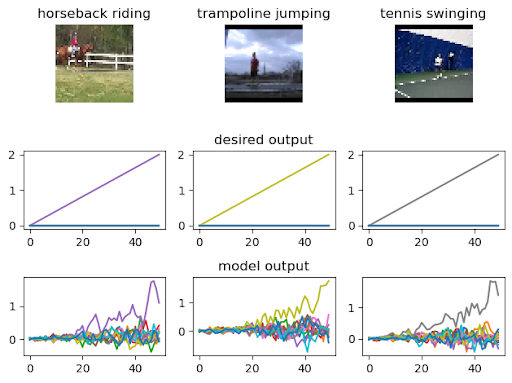}
           \caption{Sample frames of the input and their corresponding classes mentioned above, and the desired and the model predicted ramp output. Only one class ramps up linearly whereas all other classes remain zero throughout. }
           \label{fig:UCF}
        \end{figure}
           
        \subsection{Video Frame Prediction Task}

        In the video frame prediction task, the network has to predict the next frame at time $t +1$, given the input frame at time $t$ \cite{chang2021mau}. We generated a synthetic dataset to conduct the experiment. Each video in the dataset is of $40\times40$ greyscale images and the number of frames is 16. Squares can move in any one of the possible directions in 360 degrees, where dimension of each square is in the range of 2 to 4 (Figure 11). A total of 1000 such videos were generated and 8:2 train-validation split is used. First 15 frames ($t$ = 0 to $t$ = 14) is given as the input to the network, and the network should predict the next 15 frames ($t$ = 1 to $t$ = 15).

        \begin{table}[h]
            \caption{Video Frame Prediction}
            \centering
            \begin{tabular}{@{}cc@{}}
                \toprule
                Initial frequency range of oscillators & {[}0.1-7 Hz{]}\\ Input type to oscillator & I(t)                                                                                                             \\ \midrule
                Architecture                           & \begin{tabular}[c]{@{}c@{}}2 x OCNN (3x3,40 filters), \\ flatten, \\  output (2)\end{tabular} \\ \midrule
                Frequency of oscillators               & Trained                                                                                                                     \\ \bottomrule
            \end{tabular}
            \label{table:video_frame}
        \end{table}

        The network architecture is Table-\ref{table:video_frame}. Adam optimiser with a learning rate of 0.001 is used to optimise the objective function and MSE was used as the loss function. The total trainable parameters are 30,401. The validation MSE loss after training is 0.04 (p$<$0.05,n=10). A sample prediction of our model is given in Fig. \ref{fig:box}. 

        \begin{figure}[h]
            \centering
            \includegraphics[height = 4cm, width = 8cm]{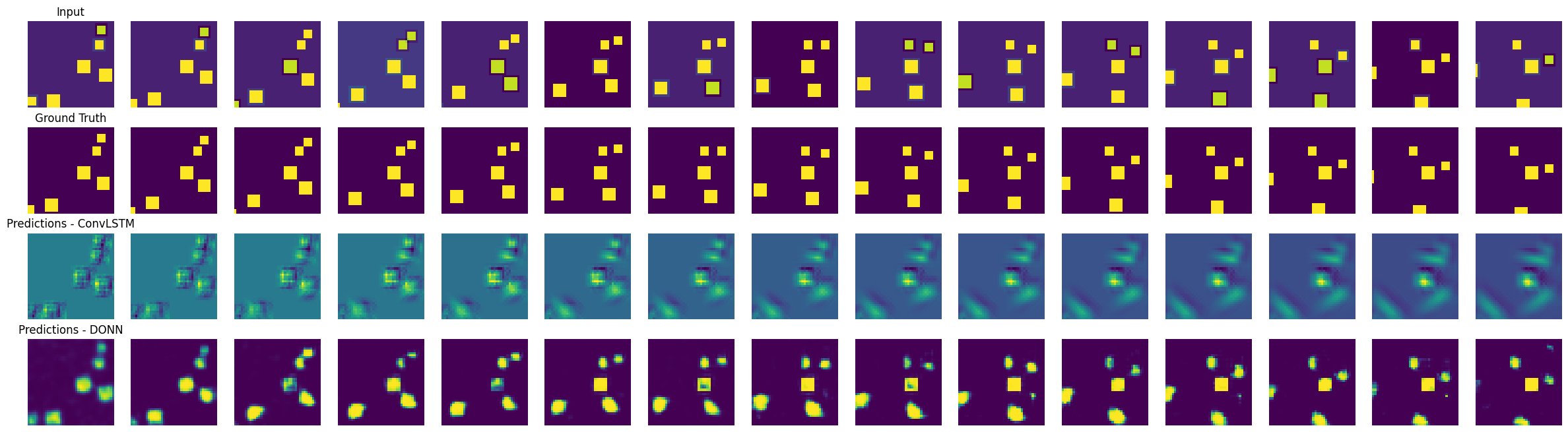}
            \caption{The row wise description is the model input, desired output, output from an equivalent ConvLSTM layer, and the prediction made by our network. The prediction by the Oscillator network visually look better.}
            \label{fig:box}
        \end{figure}
\end{section}

\begin{section}{Discussions and Concluding Remarks}

We present a broad class of trainable deep oscillatory neural networks can be trained on a variety of classification problems. We also presented an oscillatory analog of CNNs known as OCNNs that can be used to classify videos. We argue that the proposed class of DONNs is able to overcome some longstanding limitations of the existing oscillatory network models in computational neuroscience   and in artificial neural networks. 

In computational neuroscience modelling, oscillatory neural networks have been used in several subdomains of neuroscience. For instance, rhythmic movements underlying locomotion are thought to be driven by special neural circuits known as Central Pattern Generators (CPGs), capable of generating oscillatory dynamics \cite{alexander1984gaits, maufroy2008towards, grillner1996neural, ivashko2003modeling}. Thus oscillatory networks have been used to model the rhythmic movements of locomotion in bipeds or quadrupeds \cite{dutra2003modeling, bonagiri2022coupled}, or  swimming movements \cite{cao2013applying}, and rhythmic movements of a robotic hand \cite{lakatos2014nonlinear}. Some early models of locomotion used fixed connections, with bio-inspired architecture, but later models like those from Ijspeert and colleagues \cite{ijspeert2008central} are trainable models. However, this latter class of models are purely generative models i.e. they are not I/O models and generate an output without any input.

When it comes to applications of oscillatory networks in sensory processing, an important class of models is dedicated to feature binding.  The problem of feature binding addresses the question of how the brain binds the different sensory characteristics like color, form, texture to construct the presentation of a single object. It has been proposed that such binding can be achieved temporally by synchronizing the oscillations underlying various sensory characteristics, a mechanism known as temporal feature binding \cite{gray1989oscillatory}. A number of oscillatory neural networks have been proposed to describe temporal feature binding in perception \cite{isokawa2005perceptual}, multisensory integration \cite{rao2018oscillatory}, working memory \cite{pina2018oscillations}. Oscillatory neural networks are naturally suited to modelling auditory processing of music, thanks to the intrinsic rhythmicity of the signal, and have therefore been applied to music perception \cite{kimandlarge, chou2019learning}. The aforementioned class of oscillatory neural networks either have fixed connections or are trained by unsupervised learning in cases where learning occurs.

Then there is a rich body of modelling literature that applies oscillatory neural networks to model brain dynamics at large scale. These models are calibrated by various types of functional readouts of the brain like eg fMRI \cite{sip2023characterization, jirsa2022virtual}, Electroencephalogram (EEG) \cite{ghosh2023modeling, sanz2013virtual, TVB_1, schirner2018inferring}, or Magnetoencephalogram (MEG) \cite{schafer2014oscillations}.  In addition to describing the normal brain, this class of models have been applied to model brain dynamics in conditions of brain disorders including Alzheimers \cite{azh_fmri}, Stroke \cite{stroke_fmri} , Parkinson’s disease \cite{pd_fmri} etc. In this class of models, typically, the network connections are not a result of learning but are obtained from the empirical structural connectivity of the brain. However, a recent modelled study had proposed use of complex-valued connections in which the magnitudes are obtained from structural connectivity while the phase is learnt by a modified form of Hebbian learning \cite{bandyopadhyay2023phenomenological}. However, note that the aforementioned models of brain dynamics are purely generative models and not I/O models.

Apart from the aforementioned biologically oriented examples, there have been a range of artificial engineering applications of oscillatory neural networks. An important class refers to associative memory models. Earliest associative memory models are based on fixed point dynamics \cite{hopfield1982neural, hopfield1984neurons}, which have low biological plausibility are artificial since memories in the brain are necessarily composed of neural oscillations \cite{freeman1975mass, freeman2000analysis}. This motivation had inspired a range of oscillatory associative memories \cite{imai2023associative, chakravarthy1996complex}. In the earliest models of oscillatory associative memories all the oscillators had the same intrinsic frequency. Deficiencies in storage capacity shown by oscillatory associative memories that operate at a single frequency are overcome by generalizations that involve an addition of harmonics to the dynamics, apart from the fundamental frequency \cite{aonishi, nishikawa, follman}   . Note that in this entire class of oscillatory neural networks are trained by unsupervised learning, more specifically a variation  of Hebbian learning. Other artificial applications of oscillatory neural networks include image segmentation, auditory scene segmentation and object detection. \cite{wang1995locally, kuzmina2004oscillatory, kuzmina2005oscillatory}. Note that the class of oscillatory networks described in this para involve an unsupervised learning rules like  Hebbian learning or its variations and therefore cannot learn I/O behavior.

Recent years saw a surge of interest in the area of hardware implementations of oscillatory neural networks \cite{csaba2020noise}. In the 1950s, von Neumann had proposed that oscillators can be used as logic gates and digital information can be stored in the form of phase differences \cite{von2000computer, wigington1959new}. Hardware implementations of neural networks, often called neuromorphic systems, often use spiking neuron networks \cite{rathi2023exploring}. The recent interest in oscillatory neurons, as a substitute for spiking neurons, can be linked to the opportunities that oscillatory hardware offers to low-power design. Oscillatory neural hardware implementations encompass associative memories \cite{nikonov2015coupled}, image and audio segmentation \cite{rudner2024design, abernot2021digital}, memristor models of locomotor rhythms \cite{bonagiri2022coupled}. Thus the hardware implementations of the above oscillatory networks also cannot learn I/O behavior. Thus the oscillatory neural network models reviewed above involve either fixed connections or unsupervised learning. It is therefore worthwhile to formulate oscillatory neural networks that can learn I/O relations by supervised learning. The broad class of DONNs described in this paper, which to our knowledge is perhaps the first demonstration of a deep oscillatory neural network with multiple frequencies, trainable in an I/O fashion, is expected to be a valuable addition to oscillatory neural network literature.
\end{section}

\bibliography{main}

\begin{thebibliography}{10}
\providecommand{\url}[1]{#1}
\csname url@samestyle\endcsname
\providecommand{\newblock}{\relax}
\providecommand{\bibinfo}[2]{#2}
\providecommand{\BIBentrySTDinterwordspacing}{\spaceskip=0pt\relax}
\providecommand{\BIBentryALTinterwordstretchfactor}{4}
\providecommand{\BIBentryALTinterwordspacing}{\spaceskip=\fontdimen2\font plus
\BIBentryALTinterwordstretchfactor\fontdimen3\font minus \fontdimen4\font\relax}
\providecommand{\BIBforeignlanguage}[2]{{%
\expandafter\ifx\csname l@#1\endcsname\relax
\typeout{** WARNING: IEEEtran.bst: No hyphenation pattern has been}%
\typeout{** loaded for the language `#1'. Using the pattern for}%
\typeout{** the default language instead.}%
\else
\language=\csname l@#1\endcsname
\fi
#2}}
\providecommand{\BIBdecl}{\relax}
\BIBdecl

\bibitem{KELL2018630}
A.~J. Kell, D.~L. Yamins, E.~N. Shook, S.~V. Norman-Haignere, and J.~H. McDermott, ``A task-optimized neural network replicates human auditory behavior, predicts brain responses, and reveals a cortical processing hierarchy,'' \emph{Neuron}, vol.~98, no.~3, pp. 630--644.e16, 2018.

\bibitem{huang2019braininspired}
Y.~Huang, S.~Dai, T.~Nguyen, P.~Bao, D.~Y. Tsao, R.~G. Baraniuk, and A.~Anandkumar, ``Brain-inspired robust vision using convolutional neural networks with feedback,'' in \emph{Real Neurons {\&} Hidden Units: Future directions at the intersection of neuroscience and artificial intelligence @ NeurIPS 2019}, 2019.

\bibitem{wardle2020rapid}
S.~G. Wardle, J.~Taubert, L.~Teichmann, and C.~I. Baker, ``Rapid and dynamic processing of face pareidolia in the human brain,'' \emph{Nature communications}, vol.~11, no.~1, p. 4518, 2020.

\bibitem{wardle2020recent}
S.~G. Wardle and C.~I. Baker, ``Recent advances in understanding object recognition in the human brain: deep neural networks, temporal dynamics, and context,'' \emph{F1000Research}, vol.~9, 2020.

\bibitem{geirhos2022imagenettrained}
R.~Geirhos, P.~Rubisch, C.~Michaelis, M.~Bethge, F.~A. Wichmann, and W.~Brendel, ``Imagenet-trained cnns are biased towards texture; increasing shape bias improves accuracy and robustness,'' 2022.

\bibitem{cichy2016comparison}
R.~M. Cichy, A.~Khosla, D.~Pantazis, A.~Torralba, and A.~Oliva, ``Comparison of deep neural networks to spatio-temporal cortical dynamics of human visual object recognition reveals hierarchical correspondence,'' \emph{Scientific reports}, vol.~6, no.~1, p. 27755, 2016.

\bibitem{bick2020understanding}
C.~Bick, M.~Goodfellow, C.~R. Laing, and E.~A. Martens, ``Understanding the dynamics of biological and neural oscillator networks through exact mean-field reductions: a review,'' \emph{The Journal of Mathematical Neuroscience}, vol.~10, no.~1, p.~9, 2020.

\bibitem{chinichian2024modeling}
N.~Chinichian, M.~Lindner, S.~Yanchuk, T.~Schwalger, E.~Sch{\"o}ll, and R.~Berner, ``Modeling brain network flexibility in networks of coupled oscillators: a feasibility study,'' \emph{Scientific Reports}, vol.~14, no.~1, p. 5713, 2024.

\bibitem{laing2017phase}
C.~R. Laing, ``Phase oscillator network models of brain dynamics,'' \emph{Computational models of brain and behavior}, pp. 505--517, 2017.

\bibitem{flipflop}
S.~Kumari, V.~Chandrasekaran, and V.~Chakravarthy, ``The flip-flop neuron: a memory efficient alternative for solving challenging sequence processing and decision-making problems,'' \emph{Neural Computing and Applications}, 2023.

\bibitem{gopalakrishnan2020sentiment}
K.~Gopalakrishnan and F.~M. Salem, ``Sentiment analysis using simplified long short-term memory recurrent neural networks,'' 2020.

\bibitem{graves2014generating}
A.~Graves, ``Generating sequences with recurrent neural networks,'' 2014.

\bibitem{TAN20205}
Z.~Tan, S.~Wang, Z.~Yang, G.~Chen, X.~Huang, M.~Sun, and Y.~Liu, ``Neural machine translation: A review of methods, resources, and tools,'' \emph{AI Open}, vol.~1, pp. 5--21, 2020.

\bibitem{shi2015convolutional}
X.~Shi, Z.~Chen, H.~Wang, D.-Y. Yeung, W.~kin Wong, and W.~chun Woo, ``Convolutional lstm network: A machine learning approach for precipitation nowcasting,'' 2015.

\bibitem{atabay2017deep}
H.~A. Atabay, ``Deep residual learning for tomato plant leaf disease identification.'' \emph{Journal of Theoretical \& Applied Information Technology}, vol.~95, no.~24, 2017.

\bibitem{rajalingam2018review}
B.~Rajalingam, R.~Priya, and R.~Scholar, ``Review of multimodality medical image fusion using combined transform techniques for clinical application,'' \emph{Intl Journal of Sci Res in Comp Sci App.}, vol.~7, no.~3, pp. 1--8, 2018.

\bibitem{music_ref}
E.~W. Large, I.~Roman, J.~C. Kim, J.~Cannon, J.~K. Pazdera, L.~J. Trainor, J.~Rinzel, and A.~Bose, ``Dynamic models for musical rhythm perception and coordination,'' \emph{Frontiers in Computational Neuroscience}, vol.~17, 2023.

\bibitem{hinton2012deep}
G.~Hinton, L.~Deng, D.~Yu, G.~E. Dahl, A.-r. Mohamed, N.~Jaitly, A.~Senior, V.~Vanhoucke, P.~Nguyen, T.~N. Sainath \emph{et~al.}, ``Deep neural networks for acoustic modeling in speech recognition: The shared views of four research groups,'' \emph{IEEE Signal processing magazine}, vol.~29, no.~6, pp. 82--97, 2012.

\bibitem{chou2019learning}
S.-Y. Chou, K.-H. Cheng, J.-S.~R. Jang, and Y.-H. Yang, ``Learning to match transient sound events using attentional similarity for few-shot sound recognition,'' in \emph{ICASSP 2019-2019 IEEE International Conference on Acoustics, Speech and Signal Processing (ICASSP)}.\hskip 1em plus 0.5em minus 0.4em\relax IEEE, 2019, pp. 26--30.

\bibitem{Kahana1669}
M.~J. Kahana, ``The cognitive correlates of human brain oscillations,'' \emph{Journal of Neuroscience}, vol.~26, no.~6, pp. 1669--1672, 2006.

\bibitem{dang2020neuronal}
T.~Dang-Vu and R.~Courtemanche, \emph{Neuronal Oscillations of Wakefulness and Sleep: Windows on Spontaneous Activity of the Brain}.\hskip 1em plus 0.5em minus 0.4em\relax Springer New York, 2020.

\bibitem{akeju2017neural}
O.~Akeju and E.~N. Brown, ``Neural oscillations demonstrate that general anesthesia and sedative states are neurophysiologically distinct from sleep,'' \emph{Current opinion in neurobiology}, vol.~44, pp. 178--185, 2017.

\bibitem{blanco2011data}
J.~A. Blanco, M.~Stead, A.~Krieger, W.~Stacey, D.~Maus, E.~Marsh, J.~Viventi, K.~H. Lee, R.~Marsh, B.~Litt \emph{et~al.}, ``Data mining neocortical high-frequency oscillations in epilepsy and controls,'' \emph{Brain}, vol. 134, no.~10, pp. 2948--2959, 2011.

\bibitem{doi:10.1126/science.1099745}
G.~Buzsáki and A.~Draguhn, ``Neuronal oscillations in cortical networks,'' \emph{Science}, vol. 304, no. 5679, pp. 1926--1929, 2004.

\bibitem{pinotsis2014neural}
D.~Pinotsis, P.~Robinson, P.~beim Graben, and K.~Friston, ``Neural masses and fields: modeling the dynamics of brain activity,'' p. 149, 2014.

\bibitem{cowan1972autoradiographic}
W.~M. Cowan, D.~Gottlieb, A.~E. Hendrickson, J.~Price, and T.~Woolsey, ``The autoradiographic demonstration of axonal connections in the central nervous system,'' \emph{Brain research}, vol.~37, no.~1, pp. 21--51, 1972.

\bibitem{nagumo1962active}
J.~Nagumo, S.~Arimoto, and S.~Yoshizawa, ``An active pulse transmission line simulating nerve axon,'' \emph{Proceedings of the IRE}, vol.~50, no.~10, pp. 2061--2070, 1962.

\bibitem{jirsa2022virtual}
V.~Jirsa, M.~M. Woodman, and L.~Domide, ``The virtual brain (tvb): Simulation environment for large-scale brain networks,'' in \emph{Encyclopedia of Computational Neuroscience}.\hskip 1em plus 0.5em minus 0.4em\relax Springer, 2022, pp. 3397--3407.

\bibitem{bandyopadhyay2023phenomenological}
A.~Bandyopadhyay, S.~Ghosh, D.~Biswas, and et~al., ``A phenomenological model of whole brain dynamics using a network of neural oscillators with power-coupling,'' \emph{Scientific Reports}, vol.~13, p. 16935, 2023.

\bibitem{cabral2011role}
J.~Cabral, E.~Hugues, O.~Sporns, and G.~Deco, ``Role of local network oscillations in resting-state functional connectivity,'' \emph{Neuroimage}, vol.~57, no.~1, pp. 130--139, 2011.

\bibitem{bey2023lesion}
P.~Bey, K.~Dhindsa, A.~Kashyap, M.~Schirner, J.~Feldheim, M.~B{\"o}nstrup, R.~Schulz, B.~Cheng, G.~Thomalla, C.~Gerloff \emph{et~al.}, ``Lesion aware automated processing pipeline for multimodal neuroimaging stroke data and the virtual brain (tvb),'' \emph{bioRxiv}, pp. 2023--08, 2023.

\bibitem{sanz2013virtual}
P.~Sanz~Leon, S.~A. Knock, M.~M. Woodman, L.~Domide, J.~Mersmann, A.~R. McIntosh, and V.~Jirsa, ``The virtual brain: a simulator of primate brain network dynamics,'' \emph{Frontiers in neuroinformatics}, vol.~7, p.~10, 2013.

\bibitem{deco2009key}
G.~Deco, V.~Jirsa, O.~Sporns, and R.~K{\"o}tter, ``Key role of coupling, delay, and noise in resting brain fluctuations,'' \emph{Proceedings of the National Academy of Sciences}, vol. 106, no.~25, pp. 10\,302--10\,307, 2009.

\bibitem{dutra2003modeling}
M.~Dutra, A.~de~Pina~Filho, and V.~Romano, ``Modeling of a bipedal locomotor using coupled nonlinear oscillators of van der pol,'' \emph{Biol Cybern}, vol.~88, pp. 286--292, 2003.

\bibitem{lakatos2014nonlinear}
D.~Lakatos, F.~Petit, and A.~Albu-Sch{\"a}ffer, ``Nonlinear oscillations for cyclic movements in human and robotic arms,'' \emph{IEEE Transactions on Robotics}, vol.~30, no.~4, pp. 865--879, 2014.

\bibitem{cao2013applying}
Y.~Cao, S.~Bi, Y.~Cai, and L.~Zhang, ``Applying coupled nonlinear oscillators to imitate swimming modes of cow-nosed rays,'' in \emph{2013 IEEE International Conference on Robotics and Biomimetics (ROBIO)}, 2013, pp. 552--557.

\bibitem{kimandlarge}
J.~C. Kim and E.~Large, ``Signal processing in periodically forced gradient frequency neural networks,'' \emph{Frontiers in Computational Neuroscience}, vol.~9, 12 2015.

\bibitem{saccade}
S.~Kumari and V.~S. Chakravarthy, ``Biologically inspired image classifier based on saccadic eye movement design for convolutional neural networks,'' \emph{Neurocomputing}, vol. 513, pp. 294--317, 2022.

\bibitem{soman}
K.~Soman, V.~Muralidharan, and V.~S. Chakravarthy, ``A model of multisensory integration and its influence on hippocampal spatial cell responses,'' \emph{IEEE Transactions on Cognitive and Developmental Systems}, vol.~10, no.~3, pp. 637--646, 2018.

\bibitem{maas-EtAl:2011:ACL-HLT2011}
A.~L. Maas, R.~E. Daly, P.~T. Pham, D.~Huang, A.~Y. Ng, and C.~Potts, ``Learning word vectors for sentiment analysis,'' in \emph{Proceedings of the 49th Annual Meeting of the Association for Computational Linguistics: Human Language Technologies}.\hskip 1em plus 0.5em minus 0.4em\relax Association for Computational Linguistics, June 2011, pp. 142--150.

\bibitem{ucf11_dataset}
J.~Liu, J.~Luo, and M.~Shah, ``Recognizing realistic actions from videos “in the wild”,'' in \emph{2009 IEEE Conference on Computer Vision and Pattern Recognition}, 2009, pp. 1996--2003.

\bibitem{chang2021mau}
Z.~Chang, X.~Zhang, S.~Wang, S.~Ma, Y.~Ye, X.~Xinguang, and W.~Gao, ``Mau: A motion-aware unit for video prediction and beyond,'' \emph{Advances in Neural Information Processing Systems}, vol.~34, pp. 26\,950--26\,962, 2021.

\bibitem{alexander1984gaits}
R.~M. Alexander, ``The gaits of bipedal and quadrupedal animals,'' \emph{International Journal of Robotics Research}, vol.~3, no.~2, pp. 49--59, Jun 1984.

\bibitem{maufroy2008towards}
C.~Maufroy, H.~Kimura, and K.~Takase, ``Towards a general neural controller for quadrupedal locomotion,'' \emph{Neural Networks}, vol.~21, no.~4, pp. 642--653, 2008.

\bibitem{grillner1996neural}
S.~Grillner, ``Neural networks for vertebrate locomotion,'' \emph{Scientific American}, vol. 274, no.~1, pp. 64--69, 1996.

\bibitem{ivashko2003modeling}
D.~G. Ivashko, B.~I. Prilutsky, S.~N. Markin, J.~K. Chapin, and I.~A. Rybak, ``Modeling the spinal cord neural circuitry controlling cat hindlimb movement during locomotion,'' \emph{Neurocomputing}, vol. 52--54, pp. 331--338, 2003.

\bibitem{bonagiri2022coupled}
A.~Bonagiri, D.~Biswas, and S.~Chakravarthy, ``Coupled memristor oscillators for neuromorphic locomotion control: Modeling and analysis,'' \emph{IEEE Transactions on Neural Networks and Learning Systems}, 2022.

\bibitem{ijspeert2008central}
A.~J. Ijspeert, ``Central pattern generators for locomotion control in animals and robots: A review,'' \emph{Neural Networks}, vol.~21, no.~4, pp. 642--653, 2008.

\bibitem{gray1989oscillatory}
C.~Gray, P.~K{\"o}nig, A.~Engel, and W.~Singer, ``Oscillatory responses in cat visual cortex exhibit inter-columnar synchronization which reflects global stimulus properties,'' \emph{Nature}, vol. 338, no. 6213, pp. 334--337, 1989.

\bibitem{isokawa2005perceptual}
T.~Isokawa, H.~Nishimura, N.~Kamiura, and N.~Matsui, ``Perceptual binding by coupled oscillatory neural network,'' in \emph{Artificial Neural Networks: Biological Inspirations – ICANN 2005}, 2005, pp. 336--344.

\bibitem{rao2018oscillatory}
A.~R. Rao, ``An oscillatory neural network model that demonstrates the benefits of multisensory learning,'' \emph{Cognitive Neurodynamics}, vol.~12, no.~5, pp. 481--499, 2018.

\bibitem{pina2018oscillations}
J.~E. Pina, M.~Bodner, and B.~Ermentrout, ``Oscillations in working memory and neural binding: A mechanism for multiple memories and their interactions,'' \emph{PLoS Comput Biol}, vol.~14, no.~11, p. e1006517, 2018.

\bibitem{sip2023characterization}
V.~Sip, M.~Hashemi, T.~Dickscheid, K.~Amunts, S.~Petkoski, and V.~Jirsa, ``Characterization of regional differences in resting-state fmri with a data-driven network model of brain dynamics,'' \emph{Science Advances}, vol.~9, no.~11, p. eabq7547, 2023.

\bibitem{ghosh2023modeling}
S.~Ghosh, D.~Biswas, S.~Vijayan, and V.~S. Chakravarthy, ``Modeling whole brain electroencephalogram (eeg) in a spatially organized oscillatory neural network,'' \emph{bioRxiv}, vol. 2023.07.16.549247, 2023.

\bibitem{TVB_1}
A.~Al-Hossenat, P.~Wen, and Y.~Li, ``Large-scale brain network model and multi-band electroencephalogram rhythm simulations,'' \emph{International Journal of Biomedical Engineering and Technology}, vol.~38, no.~4, pp. 395--409, 2022.

\bibitem{schirner2018inferring}
M.~Schirner, A.~R. McIntosh, V.~Jirsa, G.~Deco, and P.~Ritter, ``Inferring multi-scale neural mechanisms with brain network modelling,'' \emph{elife}, vol.~7, p. e28927, 2018.

\bibitem{schafer2014oscillations}
C.~B. Sch{\"a}fer, B.~R. Morgan, A.~X. Ye, M.~J. Taylor, and S.~M. Doesburg, ``Oscillations, networks, and their development: Meg connectivity changes with age,'' \emph{Human Brain Mapping}, vol.~35, no.~10, pp. 5249--5261, 2014.

\bibitem{azh_fmri}
L.~Yang, J.~Lu, D.~Li, J.~Xiang, T.~Yan, J.~Sun, and B.~Wang, ``Alzheimer’s disease: Insights from large-scale brain dynamics models,'' \emph{Brain Sciences}, vol.~13, p. 1133, 07 2023.

\bibitem{stroke_fmri}
S.~Idesis, C.~Favaretto, N.~V. Metcalf, J.~C. Griffis, G.~L. Shulman, M.~Corbetta, and G.~Deco, ``Inferring the dynamical effects of stroke lesions through whole-brain modeling,'' \emph{NeuroImage: Clinical}, vol.~36, p. 103233, 2022.

\bibitem{pd_fmri}
J.~Kim, M.~Criaud, S.~S. Cho, M.~Díez-Cirarda, A.~Mihaescu, S.~Coakeley, C.~Ghadery, M.~Valli, M.~F. Jacobs, S.~Houle, and A.~P. Strafella, ``{Abnormal intrinsic brain functional network dynamics in Parkinson’s disease},'' \emph{Brain}, vol. 140, no.~11, pp. 2955--2967, 10 2017.

\bibitem{hopfield1982neural}
J.~J. Hopfield, ``Neural networks and physical systems with emergent collective computational abilities,'' \emph{Proceedings of the National Academy of Sciences}, vol.~79, no.~8, pp. 2554--2558, 1982.

\bibitem{hopfield1984neurons}
------, ``Neurons with graded response have collective computational properties like those of two-state neurons,'' \emph{Proceedings of the National Academy of Sciences}, vol.~81, no.~10, pp. 3088--3092, 1984.

\bibitem{freeman1975mass}
W.~J. Freeman, \emph{Mass Action in the Nervous System}.\hskip 1em plus 0.5em minus 0.4em\relax Academic Press, 1975.

\bibitem{freeman2000analysis}
W.~J. Freeman and J.~M. Barrie, ``Analysis of spatial patterns of phase in neocortical gamma eegs in rabbit,'' \emph{Journal of Neurophysiology}, vol.~84, no.~3, pp. 1266--1278, 2000.

\bibitem{imai2023associative}
Y.~Imai and T.~Taniguchi, ``Associative memory by virtual oscillator network based on single spin-torque oscillator,'' \emph{Scientific Reports}, vol.~13, p. 15809, 2023.

\bibitem{chakravarthy1996complex}
S.~Chakravarthy and J.~Ghosh, ``A complex-valued associative memory for storing patterns as oscillatory states,'' \emph{Biological Cybernetics}, vol.~75, pp. 229--238, 1996.

\bibitem{aonishi}
T.~Aonishi, ``Phase transitions of an oscillator neural network with a standard hebb learning rule,'' \emph{Phys. Rev. E}, vol.~58, pp. 4865--4871, Oct 1998.

\bibitem{nishikawa}
T.~Nishikawa, F.~C. Hoppensteadt, and Y.-C. Lai, ``Oscillatory associative memory network with perfect retrieval,'' \emph{Physica D: Nonlinear Phenomena}, vol. 197, no.~1, pp. 134--148, 2004.

\bibitem{follman}
R.~Follmann, E.~E.~N. Macau, E.~Rosa, and J.~R.~C. Piqueira, ``Phase oscillatory network and visual pattern recognition,'' \emph{IEEE Transactions on Neural Networks and Learning Systems}, vol.~26, no.~7, pp. 1539--1544, 2015.

\bibitem{wang1995locally}
D.~Wang and D.~Terman, ``Locally excitatory globally inhibitory oscillator networks,'' \emph{IEEE Transactions on Neural Networks}, vol.~6, no.~1, pp. 283--286, 1995.

\bibitem{kuzmina2004oscillatory}
M.~G. Kuzmina, E.~A. Manykin, and I.~Surina, ``Oscillatory network with self-organized dynamical connections for synchronization-based image segmentation,'' \emph{Biosystems}, vol.~76, p.~43, 2004.

\bibitem{kuzmina2005oscillatory}
M.~G. Kuzmina and E.~A. Manykin, ``Oscillatory neural network for adaptive dynamical image processing,'' in \emph{Proc. CIMCA’05}, 2005, p. 301.

\bibitem{csaba2020noise}
G.~Csaba and W.~Porod, ``Noise immunity of oscillatory computing devices,'' \emph{IEEE Journal on Exploratory Solid-State Computational Devices and Circuits}, vol.~6, no.~2, pp. 164--169, 2020.

\bibitem{von2000computer}
J.~Von~Neumann, P.~M. Churchland, and K.~Von~Neumann, \emph{The Computer and the Brain}.\hskip 1em plus 0.5em minus 0.4em\relax Yale University Press, 2000.

\bibitem{wigington1959new}
R.~L. Wigington, ``A new concept in computing,'' \emph{Proceedings of the IRE}, vol.~47, no.~4, pp. 516--523, 1959.

\bibitem{rathi2023exploring}
N.~Rathi, I.~Chakraborty, A.~Kosta, A.~Sengupta, A.~Ankit, P.~Panda, and K.~Roy, ``Exploring neuromorphic computing based on spiking neural networks: Algorithms to hardware,'' \emph{ACM Computing Surveys}, vol.~55, no.~12, p. 243, 2023.

\bibitem{nikonov2015coupled}
D.~E. Nikonov and et~al., ``Coupled-oscillator associative memory array operation for pattern recognition,'' \emph{IEEE J. Exploratory Solid-State Comput. Devices Circuits}, vol.~1, pp. 85--93, 2015.

\bibitem{rudner2024design}
T.~Rudner, W.~Porod, and G.~Csaba, ``Design of oscillatory neural networks by machine learning,'' \emph{Frontiers in Neuroscience}, vol.~18, 2024.

\bibitem{abernot2021digital}
M.~Abernot, T.~Gil, M.~Jim{\'e}nez, J.~N{\'u}{\~n}ez, M.~J. Avellido, B.~Linares-Barranco, T.~Gonos, T.~Hardelin, and A.~Todri-Sanial, ``Digital implementation of oscillatory neural network for image recognition applications,'' \emph{Frontiers in Neuroscience}, vol.~15, 2021.

\end{thebibliography}

\end{document}